\documentclass[11pt]{article}
\usepackage[utf8]{inputenc}
\usepackage{booktabs}
\usepackage{multirow}
\usepackage{graphicx}
\usepackage{xcolor}
\definecolor{abscolor}{RGB}{20,28,74}
\usepackage{amsmath}
\usepackage{hyperref}
\usepackage{geometry}
\title{Can OCR-VLMs Read Devanagari?\\
\large A Stress-Test Benchmark and Post-Correction Study}

\author{Aditya Pratap Singh\\
\texttt{adipras1407@gmail.com}}
\date{2026}

\begin{document}
\maketitle

\begin{abstract}
\color{abscolor}
OCR systems, ranging from classical engines to specialised OCR vision-language
models (OCR-VLMs) and frontier multimodal LLMs, report strong results on English
and Chinese document benchmarks, yet their behaviour on Indic scripts is largely
uncharacterised. We benchmark ten systems on Devanagari (Hindi): classical
EasyOCR; open VLMs (Qwen2.5-VL-3B, Qwen3-VL-8B, olmOCR-7B); specialised OCR-VLMs
(DeepSeek-OCR, Unlimited-OCR); and frontier closed models (Gemini~2.5~Flash,
Claude~Opus~4.7, GPT-5.5, Mistral~OCR), across four synthetic degradation
conditions and $300$ real printed scans. We report four findings. First, on clean
rendered text all ten cluster within chrF++ $91$ to $98$, so synthetic text does
not separate them. Second, under degradation the specialised OCR-VLMs are the most
fragile: DeepSeek-OCR suffers rare but catastrophic repetition failures (outputs
up to $71\times$ the reference length) that wreck its corpus mean even though its
median is the best of any system, which is why we report median and
catastrophic-rate instead of the mean. Third, on real scans nine of the ten
systems collapse (EasyOCR falls from chrF++ $93.6$ to $58.3$) and the field
spreads across a $76$-point range, so synthetic renders badly overstate Devanagari
quality. Fourth, strong English OCR does not predict Indic OCR: GPT-5.5 drops to
chrF++ $58.5$ (tying classical EasyOCR) and olmOCR-7B, the model behind
olmOCR-Bench, falls to $40.5$, while the open Qwen3-VL-8B ($75.2$, runnable on a
single 24\,GB GPU) beats GPT-5.5 and approaches Mistral; Gemini and Claude lead at
$86.3$ and $82.2$. An error taxonomy separates surface errors (numerals,
punctuation) from structural ones (conjuncts, matras, nukta), and a byte-level
(ByT5) post-corrector improves a cheap engine on its own error distribution
(chrF++ $+1.2$ to $+1.5$) but does not transfer across engines. We release the
benchmark, code, and models.\footnote{\url{https://github.com/Aditya-PS-05/devanagari-ocr-benchmark}}
\end{abstract}

\section{Introduction}
The latest wave of end-to-end OCR vision-language models (OCR-VLMs), including
DeepSeek-OCR \cite{deepseekocr}, its successor DeepSeek-OCR~2, and the recently
released Unlimited-OCR \cite{unlimitedocr}, treats document parsing as
image-to-text generation with a large language decoder. These models report
state-of-the-art results on OmniDocBench, whose documents are overwhelmingly
English and Chinese. Whether those gains hold for Indic scripts is unknown.

Devanagari, the script of Hindi and several other languages, poses challenges that
do not appear in Latin and CJK text: stacked conjunct consonants
(\textit{sa\d{m}yukt\=ak\d{s}ar}), dependent vowel signs (\textit{matras}) placed
above, below, and beside a base glyph, the connecting headline
(\textit{shirorekha}), the nukta diacritic, frequent Hindi/English code-mixing,
and two numeral systems. A model that excels on Latin text may still mishandle
these.

We ask three questions. (Q1) How accurate and how robust are modern OCR-VLMs on
Devanagari under realistic image degradation? (Q2) What do they get wrong,
categorically? (Q3) Can a lightweight post-corrector recover the errors of a cheap
engine? Our contributions are:
\begin{itemize}
  \item A controlled, multi-font, multi-condition Devanagari OCR benchmark with a
  script-aware evaluation protocol (Unicode NFC normalisation; CER/WER/chrF++).
  \item A robustness analysis showing that corpus-mean error is dominated by rare
  catastrophic repetition failures, so that median together with catastrophic-rate
  is the faithful summary.
  \item A Devanagari error taxonomy that contrasts classical-OCR and VLM failure
  modes.
  \item A distribution-matched byte-level post-corrector, with a positive result
  for matched noise and a negative cross-engine transfer result.
\end{itemize}

\section{Related Work}
\textbf{End-to-end OCR-VLMs.} GOT-OCR2 \cite{got}, Nougat, and the DeepSeek-OCR
line cast OCR as long-form generation, using a high-compression visual encoder and
an LLM decoder. Unlimited-OCR \cite{unlimitedocr} replaces the decoder's attention
with Reference Sliding Window Attention to bound the KV cache for long-document
parsing, reporting a $+6$ overall gain over DeepSeek-OCR on OmniDocBench.
\textbf{Generic VLMs} such as Qwen2.5-VL \cite{qwenvl} also perform competitive
document OCR. \textbf{Document-OCR benchmarks} such as OmniDocBench
\cite{omnidocbench} and olmOCR-Bench \cite{olmocr} drive progress with
unit-test-style checks over real PDFs, but their documents are overwhelmingly
English/Latin and Chinese, and Indic scripts are essentially absent.
\textbf{Indic OCR} has historically relied on pipeline systems; to our knowledge a
large-scale evaluation of the new OCR-VLMs and frontier LLMs on Devanagari is
absent, and that is the gap this paper fills. \textbf{OCR post-correction} as
sequence-to-sequence denoising is established for Latin and historical text; we
study it for Devanagari with a byte-level model.

\section{Benchmark Construction}
\textbf{Source text.} We use the Hindi side of the FLORES test set (997
sentences), sampling the first $N{=}100$ Devanagari sentences for the main
evaluation. FLORES is held out from all training.

\textbf{Rendering.} Each sentence is rendered to a white-background image with one
of five Devanagari fonts (Droid Sans Devanagari; Lohit Devanagari; Noto Sans
Devanagari Regular/Medium/Condensed), cycled across sentences, with line wrapping
at 1400\,px width and 40\,px type.

\textbf{Degradation conditions.} From each clean image we derive three degraded
variants: \emph{blur} (Gaussian, $\sigma\!\in\![1.0,1.8]$); \emph{noise} (additive
pixel noise on $6\%$ of pixels); and \emph{low-DPI} ($0.45\times$ downscale then
upscale, bilinear). This yields $4$ conditions $\times$ $100$ images.

\textbf{Metrics.} All references and hypotheses are Unicode NFC-normalised before
scoring. We report Character Error Rate (CER), Word Error Rate (WER), and chrF++
(character $n$-gram F-score with word order 2). Because a single visual character
(\textit{ak\d{s}ara}) spans multiple code points, code-point CER understates
structural errors; we treat this as a known limitation (\S\ref{sec:limits}).

\textbf{Real printed set.} To measure the gap between synthetic and real images we
additionally evaluate on $300$ real printed-Devanagari images with transcriptions,
sampled from the Sanskrit-OCR-Typed corpus (historical typeset scans). These are
word and short-phrase level, so we use them as a real-image robustness probe
rather than a document-parsing benchmark.

\textbf{Models.} We evaluate ten systems across four families.
\emph{Classical:} EasyOCR (Hindi and English). \emph{Open VLMs:} Qwen2.5-VL-3B and
the newer Qwen3-VL-8B (generic, prompted to transcribe verbatim) and olmOCR-7B
(the model behind olmOCR-Bench). \emph{Specialised OCR-VLMs:} DeepSeek-OCR (3B,
0.5B active; ``Free OCR'') and Unlimited-OCR (3B, 0.5B active; ``document
parsing'', Gundam mode). \emph{Frontier closed (API):} Google Gemini~2.5~Flash,
Anthropic Claude~Opus~4.7, OpenAI GPT-5.5, and Mistral OCR, evaluated on the clean
and real sets (cost-bounded), while the local open and specialised models
additionally run all four degradation conditions. VLM outputs are stripped of
layout and grounding special tokens and of bounding-box coordinates before
scoring. Local inference runs on a single NVIDIA A10G (23\,GB) in bfloat16, one
model resident at a time. We also attempted PaddleOCR, GOT-OCR2, and LlamaParse:
the first two would not run reliably in our environment (a PaddlePaddle segfault
on Amazon Linux~2023 and a processor-instantiation error), and LlamaParse returned
non-Devanagari (Latin) output on every image, so we omit all three from the
quantitative tables.

\section{Results}

\subsection{Clean accuracy: everyone looks good}
On clean rendered text all ten systems score chrF++ in a narrow $91$ to $98$ band
(Table~\ref{tab:clean}). The frontier closed models lead slightly (Claude $98.0$,
Mistral $97.6$), but classical EasyOCR, the open VLMs, and the specialised
OCR-VLMs are all within a few points. Clean synthetic text does not separate the
systems, which is exactly why degradation and real data matter.

\begin{table}[h]
\centering
\caption{Clean rendered Devanagari, $N{=}100$. CER$\downarrow$ / chrF++$\uparrow$.}
\label{tab:clean}
\small
\begin{tabular}{lcc|lcc}
\toprule
Model & CER & chrF++ & Model & CER & chrF++\\
\midrule
Claude Opus 4.7 & 0.78 & \textbf{98.0} & DeepSeek-OCR & 3.58 & 93.8\\
Mistral OCR     & 1.87 & 97.6 & EasyOCR        & 2.82 & 93.6\\
GPT-5.5         & 1.22 & 96.5 & olmOCR-7B      & 3.92 & 93.4\\
Gemini 2.5 Flash& 1.25 & 96.5 & Qwen2.5-VL-3B  & 2.35 & 93.3\\
Qwen3-VL-8B     & 1.44 & 95.9 & Unlimited-OCR  & 3.20 & 91.0\\
\bottomrule
\end{tabular}
\end{table}

\subsection{Robustness under degradation}
Table~\ref{tab:main} reports corpus CER per condition for the six systems we run
locally across all four conditions. EasyOCR and the Qwen models are nearly flat,
olmOCR is stable, Unlimited-OCR degrades moderately, and DeepSeek-OCR's corpus CER
explodes to $111.8$ under blur and $51.9$ under low-DPI.

\begin{table}[h]
\centering
\caption{Corpus CER (\%, $\downarrow$) by condition, $N{=}100$. Best per column
in \textbf{bold}.}
\label{tab:main}
\small
\begin{tabular}{lcccc}
\toprule
Model & clean & blur & noise & lowdpi\\
\midrule
EasyOCR        & 2.82 & 3.05 & 2.97 & 3.07\\
Qwen2.5-VL-3B  & 2.35 & 2.54 & 2.42 & 2.54\\
Qwen3-VL-8B    & \textbf{1.44} & \textbf{1.56} & \textbf{1.32} & \textbf{1.62}\\
olmOCR-7B      & 3.92 & 3.37 & 3.23 & 3.43\\
Unlimited-OCR  & 3.20 & 6.14 & 4.17 & 4.49\\
DeepSeek-OCR   & 3.58 & 111.8 & 3.43 & 51.9\\
\bottomrule
\end{tabular}
\end{table}

\begin{figure}[h]
\centering
\includegraphics[width=0.85\linewidth]{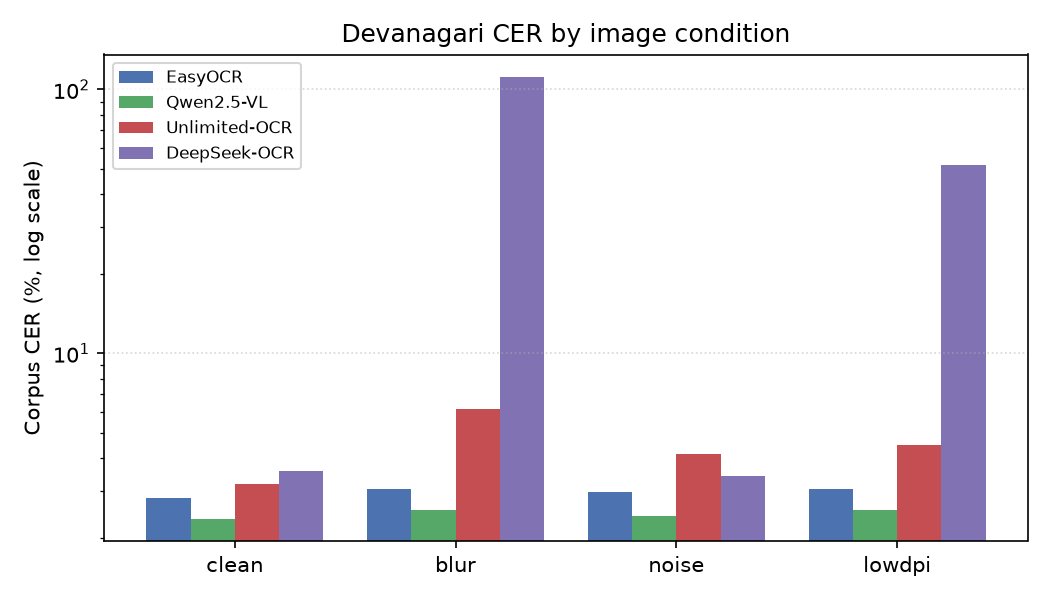}
\caption{Corpus CER by image condition (log scale). EasyOCR and Qwen are nearly
flat, while DeepSeek-OCR collapses under blur and low-DPI.}
\label{fig:cer}
\end{figure}

\textbf{The mean hides the truth.} Table~\ref{tab:robust} decomposes per-sample
CER. DeepSeek-OCR has the best median CER of all systems ($1.2$ to $1.5$), yet its
mean is wrecked by the $2$ to $3\%$ of samples that enter a degenerate repetition
loop and produce outputs up to $71.6\times$ the reference length. Unlimited-OCR,
whose decoder uses an explicit no-repeat-$n$-gram guard, bounds its worst case to
$3.8\times$. We therefore recommend reporting median CER and catastrophic rate
(the fraction of samples with CER above $50\%$) alongside the mean.

\begin{table}[h]
\centering
\caption{Per-sample CER distribution under blur and low-DPI. ``cat'' is the
fraction with CER above $50\%$; ``max$\times$'' is the largest output-to-reference
length ratio.}
\label{tab:robust}
\small
\begin{tabular}{llcccc}
\toprule
Cond. & Model & mean & median & cat\% & max$\times$\\
\midrule
\multirow{2}{*}{blur}
 & Unlimited-OCR & 6.42 & 2.40 & 1 & 3.8\\
 & DeepSeek-OCR  & 73.65 & \textbf{1.50} & 2 & 71.6\\
\midrule
\multirow{2}{*}{lowdpi}
 & Unlimited-OCR & 4.26 & 2.36 & 2 & 1.2\\
 & DeepSeek-OCR  & 58.22 & \textbf{1.54} & 3 & 34.2\\
\bottomrule
\end{tabular}
\end{table}

\begin{figure}[h]
\centering
\includegraphics[width=0.85\linewidth]{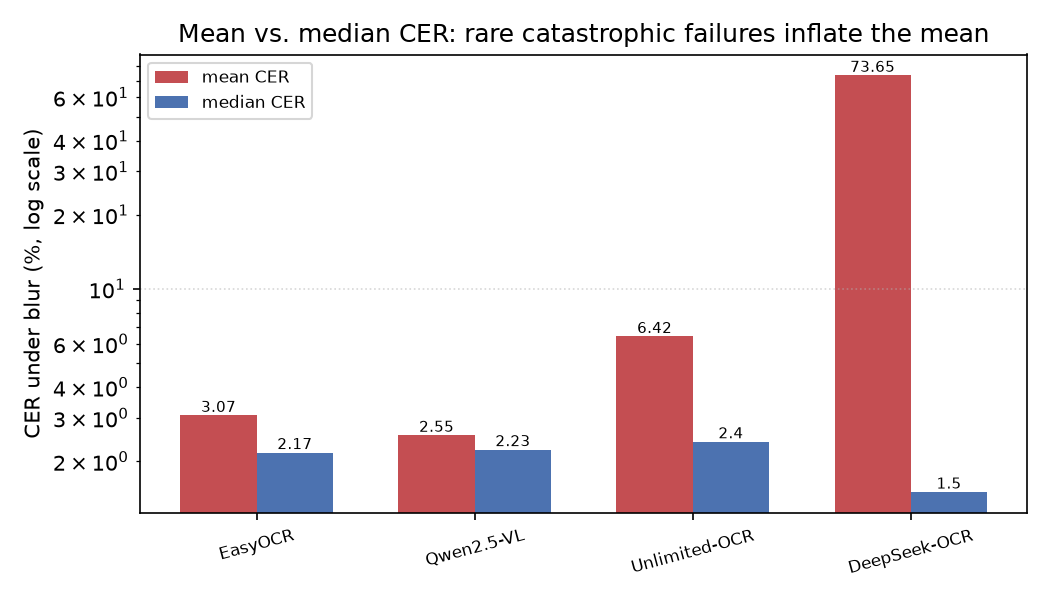}
\caption{Under blur, DeepSeek-OCR has the best median CER ($1.5$) but a
catastrophic mean ($73.7$), because $2\%$ of samples enter a repetition loop.
Median together with catastrophic rate is the faithful summary.}
\label{fig:meanmed}
\end{figure}

\textbf{The reported ordering does not transfer.} Unlimited-OCR is reported to beat
DeepSeek-OCR by $+6$ overall on Latin and CJK OmniDocBench. On clean Devanagari the
ordering reverses: DeepSeek-OCR attains higher chrF++ ($93.84$ versus $91.04$) and
lower WER. The two specialised OCR-VLMs are also both outperformed in robustness by
a generic VLM (Qwen) and by classical EasyOCR.

\subsection{Error taxonomy}
We align each hypothesis to its reference at the character level and classify every
edit into Devanagari-specific categories (Table~\ref{tab:tax}). Catastrophic
repetition samples are excluded so the taxonomy reflects genuine recognition
errors.

\begin{table}[h]
\centering
\caption{Error counts by category, clean condition, $N{=}100$.}
\label{tab:tax}
\small
\begin{tabular}{lccccccc}
\toprule
Model & conjunct & nukta & anusvara & matra & numeral & look-alike & total\\
\midrule
EasyOCR       & 3  & 4  & 7  & 13 & \textbf{69} & 73 & 197\\
Qwen2.5-VL    & 37 & 31 & 21 & 15 & 3  & 84 & 203\\
Unlimited-OCR & \textbf{68} & 48 & 30 & \textbf{44} & 4 & 45 & 255\\
DeepSeek-OCR  & 33 & \textbf{51} & 13 & 24 & 7 & 21 & 158\\
\bottomrule
\end{tabular}
\end{table}

\begin{figure}[h]
\centering
\includegraphics[width=0.85\linewidth]{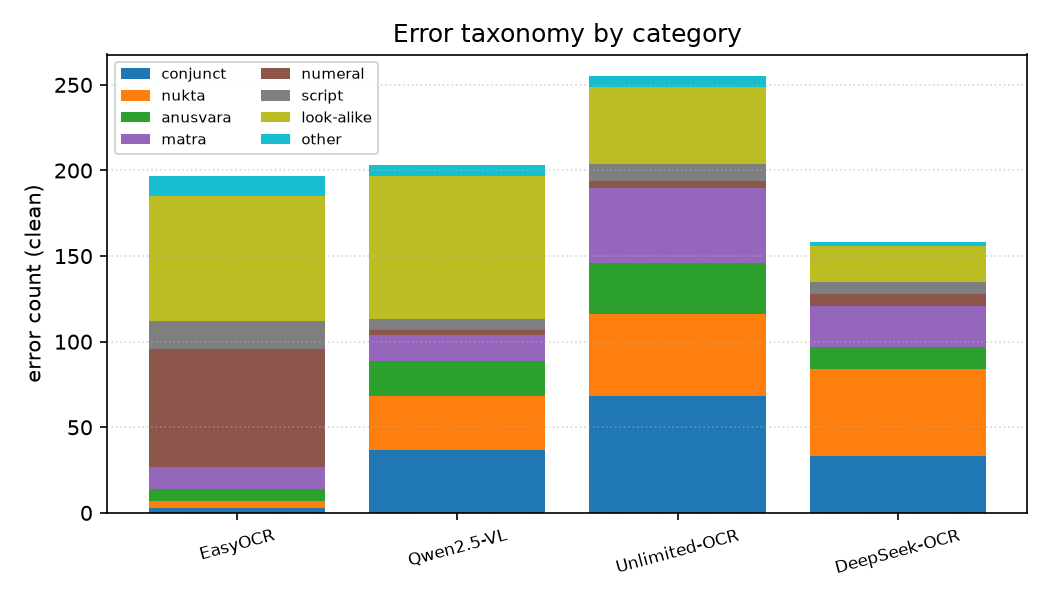}
\caption{Error composition by category (clean). EasyOCR errors are dominated by
numerals and punctuation, while the VLMs fail on structural elements (conjunct,
matra, nukta).}
\label{fig:tax}
\end{figure}

Two profiles emerge. The classical engine fails on surface elements: Devanagari
numerals (it transcribes them as Latin digits or misreads them) and punctuation
(for example, danda \textit{|} normalisation). The VLMs fail on structural
elements: conjuncts, matras, and nukta. Unlimited-OCR makes the most structural
errors, consistent with its lower chrF++. Recurring look-alike confusions are
visually and phonetically motivated and are consistent across systems:
\textit{ba}$\leftrightarrow$\textit{va}, \textit{gha}$\leftrightarrow$\textit{dha},
\textit{ma}$\leftrightarrow$\textit{bha}, \textit{da}$\leftrightarrow$\textit{dha},
and \textit{ta}$\leftrightarrow$\textit{Ta}. We also note that a substantial share
of the look-alike edits are really punctuation normalisation (danda versus
full-stop, smart quotes); such differences inflate raw error counts and should be
normalised, a methodological caveat for Indic OCR evaluation.

\subsection{Synthetic versus real printed Devanagari}
\label{sec:real}
\begin{figure}[h]
\centering
\includegraphics[width=0.85\linewidth]{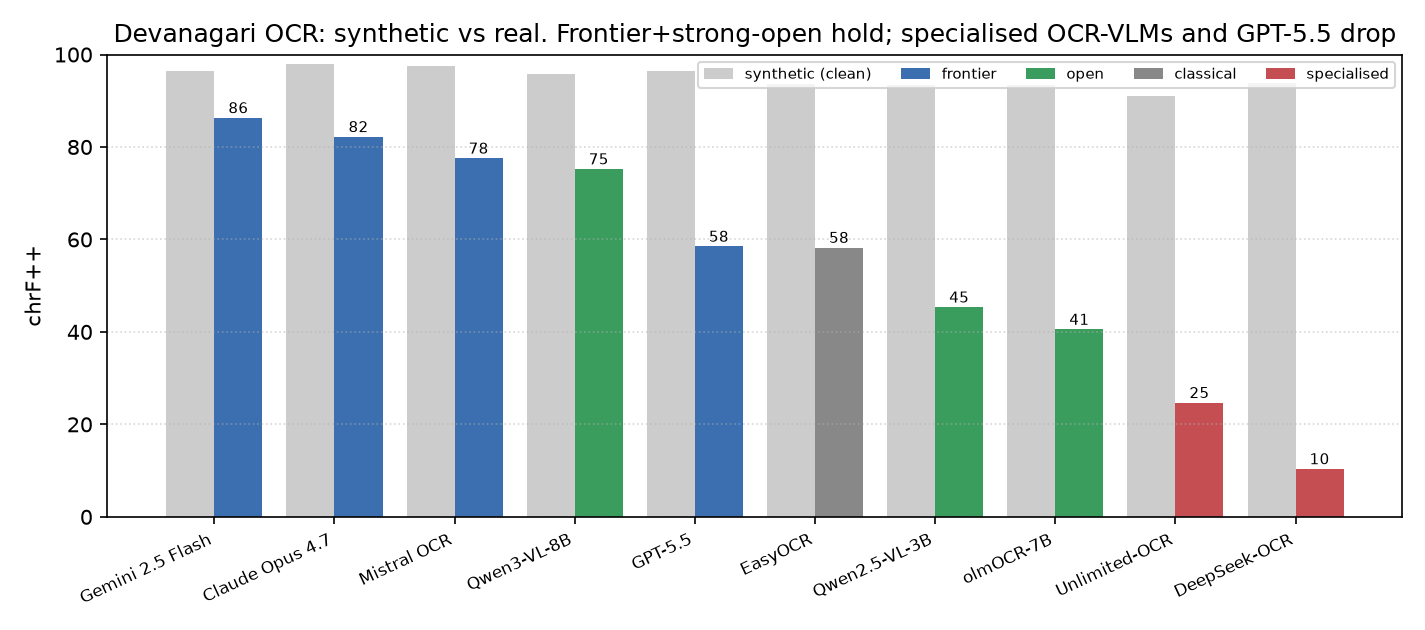}
\caption{chrF++ on synthetic clean renders versus real printed scans, ten systems
sorted by real-data chrF++ (real bars coloured by family). Nearly all collapse on
real images. Gemini, Claude, Mistral, and the open Qwen3-VL stay strong, while
GPT-5.5 and olmOCR-7B drop sharply despite strong English performance.}
\label{fig:real}
\end{figure}

The clean ties vanish on $300$ real printed-Devanagari scans
(Table~\ref{tab:real}, Fig.~\ref{fig:real}), which spread the ten systems across a
$76$-point chrF++ range. Four findings stand out.

\textbf{(1) Synthetic renders badly overstate quality.} Nine of the ten systems
drop sharply from synthetic to real; EasyOCR falls from chrF++ $93.6$ to $58.3$
and its median CER rises from about $2\%$ to $17\%$. Benchmarks built only on
rendered text are misleading for Devanagari.

\textbf{(2) Specialised OCR-VLMs collapse.} DeepSeek-OCR's median CER is $100\%$
(with $89\%$ of samples catastrophic) and Unlimited-OCR emits on average $4\times$
the reference length through hallucination; both sit at the bottom (chrF++ $10$ to
$25$).

\textbf{(3) Frontier closed models mostly hold.} Gemini, Claude, and Mistral all
reach a median CER of $0.0$ with chrF++ between $77$ and $86$. This is not a simple
``closed beats open'' story, as finding (4) shows.

\textbf{(4) The English ranking does not transfer.} Two results break the
olmOCR-Bench and English ordering. First, GPT-5.5, a top model on English document
OCR, drops to chrF++ $58.5$ on real Devanagari, tying classical EasyOCR and
falling far below Gemini and Claude. Second, the open Qwen3-VL-8B reaches chrF++
$75.2$ (median CER $0.0$), beating GPT-5.5 and approaching Mistral even though it
runs freely on a single 24\,GB GPU. Most pointedly, olmOCR-7B, the model behind the
olmOCR-Bench leaderboard, collapses to chrF++ $40.5$ on real Devanagari. Strong
English OCR performance is thus a poor predictor of Indic performance.

We acknowledge a confound: these images are word and short-phrase level, which
disadvantages page-oriented models. Even granting this, the gap between Gemini and
GPT-5.5, or between Qwen3-VL and olmOCR, occurs within the same regime and cannot
be explained by granularity alone.

\begin{table}[h]
\centering
\caption{Real printed-Devanagari scans ($N{=}300$, word-level), sorted by chrF++.
``cat'' is the CER-above-$50\%$ rate; ``len$\times$'' is the mean
output-to-reference length. F is frontier closed, O is open, C is classical, and S
is specialised OCR-VLM.}
\label{tab:real}
\small
\begin{tabular}{llccccc}
\toprule
Model & & mean CER & med CER & cat\% & len$\times$ & chrF++\\
\midrule
Gemini 2.5 Flash & F & \textbf{4.4} & \textbf{0.0} & \textbf{0.7} & 1.0 & \textbf{86.3}\\
Claude Opus 4.7  & F & 5.1 & 0.0 & 0.7 & 1.0 & 82.2\\
Mistral OCR      & F & 90.7 & 0.0 & 8.0 & 1.8 & 77.6\\
Qwen3-VL-8B      & O & 9.3 & 0.0 & 3.3 & 1.0 & 75.2\\
GPT-5.5          & F & 18.4 & 12.5 & 7.7 & 0.9 & 58.5\\
EasyOCR          & C & 34.3 & 16.7 & 26.3 & 0.7 & 58.3\\
Qwen2.5-VL-3B    & O & 26.4 & 20.3 & 14.7 & 1.0 & 45.4\\
olmOCR-7B        & O & 59.1 & 20.0 & 12.0 & 1.3 & 40.5\\
Unlimited-OCR    & S & 359.8 & 50.0 & 46.7 & 4.0 & 24.7\\
DeepSeek-OCR     & S & 112.0 & 100.0 & 89.0 & 1.1 & 10.4\\
\bottomrule
\end{tabular}
\end{table}

\section{Distribution-Matched Post-Correction}
We test whether a cheap engine can be rescued by a post-corrector that maps noisy
OCR text to clean text. We fine-tune \textbf{ByT5-small} (byte-level, which suits
character-level OCR noise) on $6{,}000$ real (OCR-output, clean) pairs. Held-out
Hindi sentences (IITB and a general corpus, disjoint from FLORES) are rendered
under the same four conditions and transcribed with EasyOCR. At inference we chunk
inputs to at most $90$ characters, because the corrector is trained on short spans
and long inputs otherwise induce repetition, then rejoin the pieces.

\begin{table}[h]
\centering
\caption{Post-correction (ByT5-small trained on EasyOCR noise). CER$\downarrow$ /
chrF++$\uparrow$, before $\rightarrow$ after.}
\label{tab:pc}
\small
\begin{tabular}{lcc}
\toprule
Target & CER (before$\rightarrow$after) & chrF++ (before$\rightarrow$after)\\
\midrule
EasyOCR clean   & 2.82 $\rightarrow$ \textbf{2.62} & 93.55 $\rightarrow$ \textbf{95.04}\\
EasyOCR noise   & 2.97 $\rightarrow$ \textbf{2.73} & 93.27 $\rightarrow$ \textbf{94.93}\\
EasyOCR blur    & 3.05 $\rightarrow$ 3.50 & 92.96 $\rightarrow$ \textbf{94.35}\\
EasyOCR lowdpi  & 3.07 $\rightarrow$ 4.53 & 93.18 $\rightarrow$ \textbf{94.10}\\
\midrule
Qwen clean      & 2.35 $\rightarrow$ 2.98 & 93.30 $\rightarrow$ 92.98\\
Unlimited clean & 3.20 $\rightarrow$ 4.93 & 91.04 $\rightarrow$ 89.77\\
DeepSeek clean  & 3.58 $\rightarrow$ 5.62 & 93.84 $\rightarrow$ 92.39\\
\bottomrule
\end{tabular}
\end{table}

\begin{figure}[h]
\centering
\includegraphics[width=0.85\linewidth]{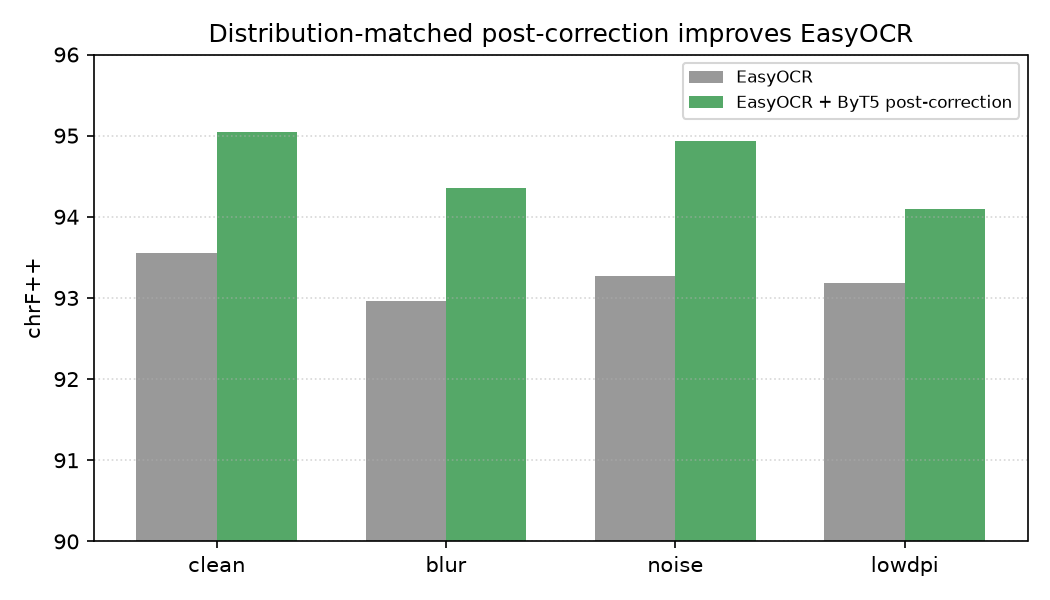}
\caption{A ByT5 post-corrector trained on EasyOCR's own error distribution
improves EasyOCR chrF++ in every condition.}
\label{fig:pc}
\end{figure}

The corrector consistently improves the engine it was trained on (EasyOCR chrF++
rises by $1.2$ to $1.5$ in all conditions, and CER improves on clean and noise).
It does not transfer: applied to Qwen, Unlimited, or DeepSeek outputs, whose error
distributions differ, it is neutral or harmful. The practical conclusion is that
OCR post-correction is effective but must be matched to the target engine's error
distribution.

\section{Limitations}
\label{sec:limits}
Our main controlled evaluation uses rendered images, whose low baseline CER limits
post-correction headroom. We partly address this with the real printed set
(\S\ref{sec:real}), but that set is word and short-phrase level and Sanskrit
typeset rather than sentence-level Hindi. An openly available, sentence-level,
real-scanned Hindi corpus with reliable transcriptions remains scarce, and
obtaining one is the most valuable next step. The controlled evaluation uses
$N{=}100$ sentences and Hindi only, CER is computed at the code-point rather than
grapheme-cluster level, and we evaluate single-page images rather than the
multi-page long-horizon setting that Unlimited-OCR targets. Line- and
document-level real Hindi data, multi-script coverage, and grapheme-aware scoring
are clear next steps.

\section{Conclusion}
Benchmarking ten OCR systems on Devanagari yields a consistent message: clean
rendered text hides every difference, and only real degraded scans reveal them.
The specialised OCR-VLMs are the least safe choice, since DeepSeek-OCR's strong
median is masked by catastrophic repetition and both it and Unlimited-OCR collapse
on real scans. Crucially, strong English OCR does not predict Indic OCR: GPT-5.5
and olmOCR-7B, both strong on English document benchmarks, fall to the middle and
bottom of the real-Devanagari ranking, while the open Qwen3-VL-8B, runnable on a
single 24\,GB GPU, beats GPT-5.5 and trails only the strongest closed models
(Gemini and Claude). Synthetic-only Devanagari benchmarks are therefore
misleading, real evaluation is indispensable, and the best general OCR for English
may be far from the best for Indic. We also contribute a robustness methodology
(median and catastrophic-rate over the mean), a Devanagari error taxonomy, and a
distribution-matched post-corrector. We release the benchmark and code to support
evaluation of OCR systems on Indic scripts.

\end{document}